\documentclass{article}

\usepackage{arxiv}

\usepackage[utf8]{inputenc} 
\usepackage[T1]{fontenc}    
\usepackage{hyperref}       
\usepackage{url}            
\usepackage{booktabs}       
\usepackage{amsfonts}       
\usepackage{nicefrac}       
\usepackage{microtype}      
\usepackage{cleveref}       
\usepackage{lipsum}         
\usepackage{graphicx}
\usepackage{natbib}
\usepackage{doi}
\usepackage{tabularx}

\title{Predicting High-Flow Nasal Cannula Failure in an ICU Using a Recurrent Neural Network with Transfer Learning and Input Data Perseveration: A Retrospective Analysis}

\date{}

\author{ George A. Pappy \\
	Virtual Pediatric Intensive Care Unit\\
	Children's Hospital Los Angeles\\
	Los Angeles, CA 90027\\
	\texttt{gpappy@chla.usc.edu} \\
	\And
	Melissa D. Aczon \\
	Virtual Pediatric Intensive Care Unit\\
	Children's Hospital Los Angeles\\
	Los Angeles, CA 90027 \\
	\texttt{maczon@chla.usc.edu} \\
	\And
	Randall C. Wetzel \\
	Virtual Pediatric Intensive Care Unit\\
	Children's Hospital Los Angeles\\
	Los Angeles, CA 90027\\
	\texttt{rwetzel@chla.usc.edu} \\
	\And
	\href{https://orcid.org/0000-0003-0382-5086
}{\includegraphics[scale=0.06]{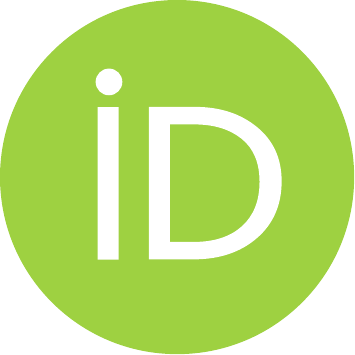}\hspace{1mm}David~R.~Ledbetter} \\
	Virtual Pediatric Intensive Care Unit\\
	Children's Hospital Los Angeles\\
	Los Angeles, CA 90027 \\
	\texttt{dledbetter@chla.usc.edu} \\
}

\renewcommand{\shorttitle}{Predicting HFNC Failure Using an RNN}

\hypersetup{
pdftitle={Predicting HFNC Failure Using an RNN},
pdfsubject={},
pdfauthor={George A. Pappy, Melissa D. Aczon, Randall C. Wetzel, David R. Ledbetter},
pdfkeywords={High Flow Nasal Cannula, HFNC Failure, Deep Learning, Recurrent Neural Network, RNN, Long Short-Term Memory, LSTM, Transfer Learning, Input Data Perseveration, Machine Learning, Critical Care},
}

\begin{document}
\maketitle

\begin{abstract}

\textbf{Background:} High Flow Nasal Cannula (HFNC) provides non-invasive respiratory support for critically ill children who may tolerate it more readily than other Non-Invasive (NIV) techniques such as Bilevel Positive Airway Pressure (BiPAP) and Continuous Positive Airway Pressure (CPAP). Moreover, HFNC may preclude the need for mechanical ventilation (intubation). Nevertheless, NIV or intubation may ultimately be necessary for certain patients. Timely prediction of HFNC failure can provide an indication for increasing respiratory support.

\textbf{Objective:} This work developed and compared machine learning models to predict HFNC failure.

\textbf{Methods:} A retrospective study was conducted using the Virtual Pediatric Intensive Care Unit database of Electronic Medical Records (EMR) of patients admitted to a tertiary pediatric ICU from January 2010 to February 2020. Patients <19 years old, without apnea, and receiving HFNC treatment were included. A Long Short-Term Memory (LSTM) model using 517 variables (vital signs, laboratory data and other clinical parameters) was trained to generate a continuous prediction of HFNC failure, defined as escalation to NIV or intubation within 24 hours of HFNC initiation. For comparison, seven other models were trained: a Logistic Regression (LR) using the same 517 variables, another LR using only 14 variables, and five additional LSTM-based models using the same 517 variables as the first LSTM and incorporating additional ML techniques (transfer learning, input perseveration, and ensembling). Performance was assessed using the area under the receiver operating curve (AUROC) at various times following HFNC initiation. The sensitivity, specificity, positive and negative predictive values (PPV, NPV) of predictions at two hours after HFNC initiation were also evaluated. These metrics were also computed in a cohort with primarily respiratory diagnoses.

\textbf{Results:} 834 HFNC trials [455 training, 173 validation, 206 test] met the inclusion criteria, of which 175 [103, 30, 42] (21.0\%) escalated to NIV or intubation. The LSTM models trained with transfer learning generally performed better than the LR models, with the best LSTM model achieving an AUROC of 0.78, vs 0.66 for the 14-variable LR and 0.71 for the 517-variable LR, two hours after initiation. All models except for the 14-variable LR achieved higher AUROCs in the respiratory cohort than in the general ICU population.

\textbf{Conclusions:}  Machine learning models trained using EMR data were able to identify children at risk for failing HFNC within 24 hours of initiation. LSTM models that incorporated transfer learning, input data perseveration and ensembling showed improved performance than the LR and standard LSTM models.

\end{abstract}


\section{Introduction}

\subsection{Background}
The use of High Flow Nasal Cannula (HFNC) respiratory support in children in critical care, emergency departments, and general wards has increased in recent years [1-8]. HFNC provides an alternative to other NIV techniques and endotracheal intubation, has fewer associated risks and complications, and children tolerate it well [3,5,6,8]. Nevertheless, many patients require escalation to a higher level of respiratory support [3,8]. Importantly, for those who require escalation, recent research indicates better clinical outcomes for patients who were escalated to higher levels of respiratory support earlier: lower hospital and ICU mortality rates, higher extubation success rate, higher ventilator-free days, and lower hospital and ICU lengths of stay [9,10].  These findings suggest that early identification of children in whom HFNC will not be successful could allow more timely institution of advanced respiratory support and decrease morbidity and mortality.

\subsection{Goals}
This study aims to develop a model to make reliable, real time predictions of a child’s response to HFNC. Such a model could help clinicians differentiate three groups: a) children likely to do well on HFNC alone, b) children likely to need higher-level support, and c) children whose HFNC response is unclear. Differentiating these three groups would help clinicians resolve the dilemma of appropriate non-invasive ventilation while not unduly and potentially harmfully prolonging it. The last group may benefit from the closest and most frequent monitoring. The second group, while still monitored frequently, could be escalated by clinicians to a higher level of support earlier. A further goal is to compare different algorithms, from logistic regressions to Long Short-Term Memory (LSTM)-based recurrent neural networks, for predicting HFNC response. Other techniques, such as transfer learning, input data perseveration, and ensembling are also explored and evaluated for their impact on performance when used with LSTMs. 

\subsection{Related Prior Work}
The authors are unaware of any studies developing a predictive model  of HFNC failure, although a few studies have investigated risk factors for escalation from HFNC to a higher level of respiratory support. Guillot, et al found that high pCO2 was a risk factor for HFNC failure in children with bronchiolitis [11]. Er, et al reported that respiratory acidosis, low initial SpO2 and SF ratio, and SF ratio < 195 during the first few hours were associated with unresponsiveness to HFNC in children with severe bacterial pneumonia in a pediatric emergency department [12]. In a small study of children with bacterial pneumonia, Yurtseven and Saz saw higher failure rates in those with higher respiratory rates [13].  Kelley et. al found that high respiratory rate, high initial venous PCO2, and pH less than 7.3 were associated with failure of HFNC [8]. 

\section{Methods}

\subsection{Data Sources}
Data for this study came from de-identified clinical observations collected in Electronic Medical Records (EMR, Cerner) of children admitted to the Pediatric Intensive Care Unit (PICU) of Children’s Hospital Los Angeles (CHLA) between January 2010 and February 2020. An episode represents a single admission and contiguous stay in the PICU. Patients may have more than one episode. EMR data for an episode included irregularly, sparsely and asynchronously charted physiologic measurements (e.g., heart rate, blood pressure), laboratory results (e.g., creatinine, glucose level), drugs (e.g., epinephrine, furosemide) and interventions (e.g., intubation, BiPAP or HFNC). Data previously collected for Virtual Pediatric Services, LLC participation [14], including diagnoses, gender, race, and disposition at discharge, were linked with the EMR data before de-identification. The CHLA Institutional Review Board (IRB) reviewed the study protocol and waived the requirement for consent and IRB approval.   

\subsection{Definitions}
For ease of reference, Table 1 lists the terminologies and definitions used throughout the sections which follow.

\begin{table}[]
\caption{Useful definitions.}
\label{tab:HFNC_Defs}
\centering
\begin{tabularx}{\linewidth}{l|l}
\hline
\textbf{Term}   & \multicolumn{1}{c|}{\textbf{Definition}}                                                                                                                                              \\ \hline
Episode         & An individual child’s single, contiguous stay in the Pediatric ICU, spanning the time between \\ & admission and discharge.                                                            \\ \hline
HFNC initiation & The start of HFNC treatment for a child not currently on HFNC. \\ &                                                                                                                      \\ \hline
HFNC period     & The 24 hours following an HFNC initiation where the child was not on HFNC support at any \\ & time during the   preceding 24 hours.                                                      \\ \hline
HFNC trial      & An episode or subset of an episode (starting with admission) where only the very last HFNC \\ & period is designated as the training target. It may include previous HFNC initiations. \\ \hline
\end{tabularx}
\end{table}

In an episode where HFNC is initiated only once, there is exactly one HFNC period and one HFNC trial. Episodes can have multiple HFNC initiations. In such cases, a single episode may have multiple HFNC periods, and each has an associated HFNC trial. Note that not all HFNC initiations have a corresponding HFNC period. For instance, if a child started on HFNC for the first time during an episode, then this marked the start of an HFNC period. If HFNC was withdrawn two hours later, and the child again received HFNC an hour after that, then this new HFNC initiation did not mark the start of a new HFNC period because the original HFNC period had not yet ended. On the other hand, if this second HFNC initiation took place > 24 hours after the first HFNC was stopped, then this second initiation marked the start of a new HFNC period because it was initiated after the first HFNC period had already ended.  Finally, at least 30 minutes was required between any de-escalation (“step-down”) from NIV or intubation prior to the start of a HFNC period. This rule was necessary since patients on a higher level of support may be stepped down to HFNC to assess their ability to breathe on their own. If such breathing trials fail, a not uncommon occurrence, these patients immediately escalate back to mechanical ventilation or NIV, technically becoming HFNC failures, but were in fact extubation, etc. failures and are not representative of the escalation scenarios of interest in this study. Figure 1 illustrates these terminologies.

\subsection{Data Inclusions and Exclusions}
Only episodes in which HFNC was used were included. Episodes of patients 19 years or older at admission were excluded, as were episodes associated with sleep apnea. Any episode which ended less than 24 hours into an HFNC period where the patient next went to the operating room was also excluded. Episodes with a Do Not Intubate (DNI) or Do Not Resuscitate (DNR) order were also excluded. 

\begin{figure}[!t]
    \centering
    \includegraphics[width=.8\linewidth]{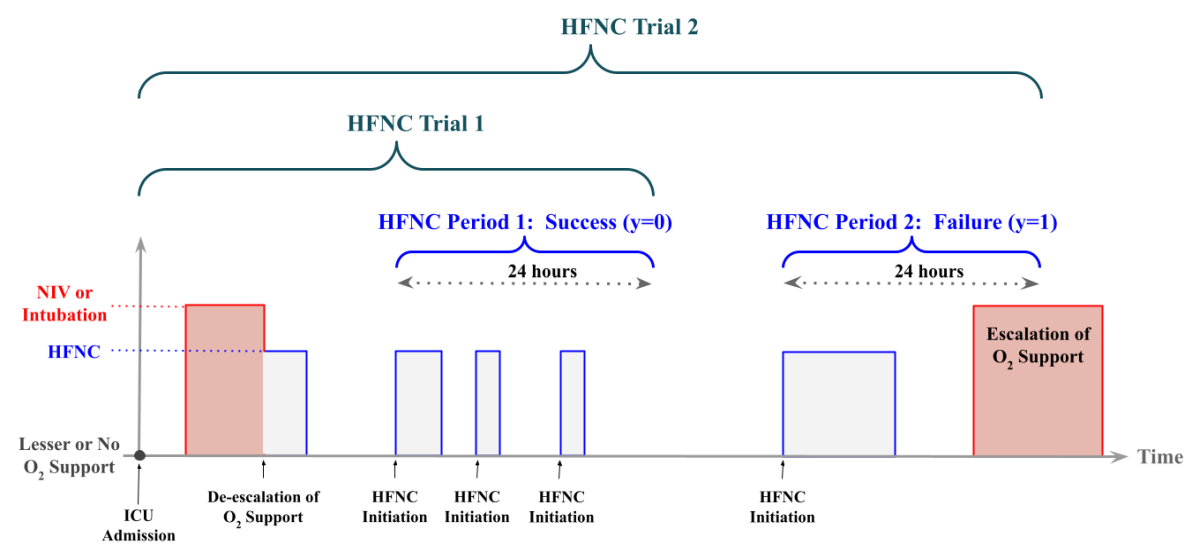}
    \caption{Illustration of HFNC scenarios, definitions, and outcomes.}
    \label{fig:Scenarios}
\end{figure}

\subsection{Target Outcome}
For each HFNC trial, the target of interest was escalation to a higher level of support (BIPAP, NIMV, intubation) within the 24 hour window (HFNC period) after HFNC initiation.  Each HFNC trial was labeled either a failure (if there was an escalation within the associated HFNC period) or a success (if there was no such escalation within the associated HFNC period).

When a patient was discharged from the PICU within 24 hours after HFNC initiation, the target label was determined by the patient’s disposition at discharge (Table 2). Episodes with the dispositions ‘Operating Room’,  ‘Another Hospital's ICU’, or ‘Another ICU in Current Hospital’ were excluded since the outcome  was ambiguous. HFNC trials associated with a favourable disposition (‘General Care Floor’, ‘Home’, or ‘Step-Down Unit’) during the HFNC period were labeled as success.

\begin{table}[]
\caption{Target outcome mappings for HFNC periods cut short by patient discharge.}
\label{tab:HFNC_Outcomes}
\centering
\begin{tabular}{|l|l|}
\hline
\textbf{Episode Disposition} & \textbf{Target Outcome Mapping} \\ \hline
\begin{tabular}[c]{@{}l@{}}General Care Floor  \\ Home  \\ Step-Down Unit/Intermediate Care   Unit\end{tabular} & Success \\ \hline
\begin{tabular}[c]{@{}l@{}}Operating Room \\ Another Hospital's ICU  \\ Another ICU in Current Hospital\end{tabular} & Censored \\ \hline
\end{tabular}
\end{table}

The HFNC definitions and outcome, combined with the exclusion criteria, resulted in 834 HFNC trials which were randomly divided into training, validation and hold-out test sets. All HFNC trials of an individual patient were in only one of these three sets to prevent leakage and bias during model evaluations. No additional stratifications were applied.

\subsection{Labeling Time Series Data for Model Training and Assessment}
Recall that the task is to predict HFNC failure (escalation of care) for each HFNC trial. Data processing starts at the beginning of the HFNC trial, with a model trained to output a prediction each time a new measurement becomes available. Figure 2 illustrates how the time series data of each HFNC trial was labelled for this process. All prediction times during the HFNC period are labeled either 1 (failure) or 0 (success).  Predictions at times before the HFNC Period are labeled NaN (“Not a Number”) to exclude the predictions here from error metrics during model training and performance evaluations.  

\subsection{Data Preprocessing}
Each episode’s time series data was converted into a matrix. Rows contain the measurements (recorded or imputed) of all variables at one time point, and columns contain values of a single variable at different times. The steps of this conversion are described in detail in previous work [15] and consists of aggregation and normalization of observed measurements, followed by imputation of missing data. A brief description is provided here:

\begin{figure}[!t]
    \centering
    \includegraphics[width=.8\linewidth]{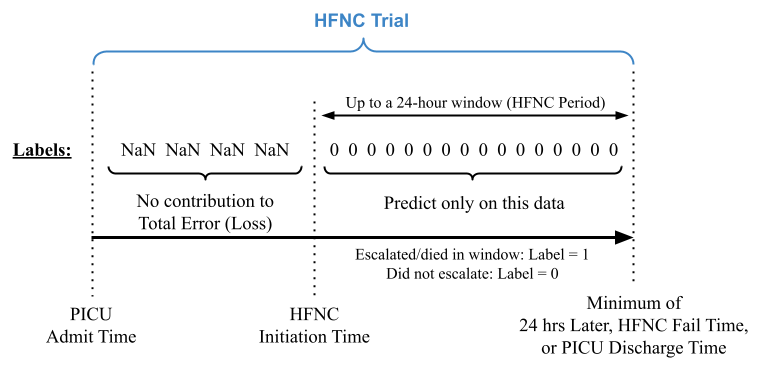}
    \caption{Illustration of labeling time series data for predicting HFNC escalation.}
    \label{fig:Scenarios}
\end{figure}

\subsubsection{Aggregation and Normalization}
Where medically appropriate, values of the same variable obtained using independent measurement methods were aggregated into a single feature. For example, invasive and non-invasive systolic blood pressure (SBP) measurements were grouped into a single variable representing SPB [16]. Any drug or intervention administered in less than 1\% of patient episodes in the training set was excluded. This aggregation and exclusion process resulted in a list of 516 distinct demographic, physiologic, laboratory, and therapy variables available as model inputs (see Tables A-1 to A-4 in Multimedia Appendix 1 for the full list; variable acronyms appear in Table A-5). Measurements considered incompatible with human life were filtered out using established minimum and maximum acceptable values (e.g., heart rates exceeding 400 beats per minute). Physiologic variables and laboratory measurements were transformed to have zero mean and unit variance using means and standard deviations derived from the training set. Administered patient therapies were scaled to the interval [0,1] using clinically defined upper limits. No variables were normalized by age because patient age was one of the inputs. Diagnoses were only used for descriptive analyses and not as model input features.

\subsubsection{Imputation}
EMR measurements were sparsely, asynchronously, and irregularly charted, with time between measurements ranging from one minute to several hours. At any time where at least one variable had a recorded value, the missing values for other unrecorded variables were imputed. The imputation process depended on the type of variable. Missing measurements for a drug or an intervention variable were set to zero, indicating absence of treatment. When a physiologic observation or lab measurement was available, it was propagated forward until another measurement was recorded. This choice reflects clinical practice and is based on the observation that measurements are recorded more frequently when the patient is unstable and less frequently when the patient appears stable [17]. If a physiologic or laboratory variable had no recorded value throughout the entire episode, its mean from the training set population was used.

\subsection{Input Perseveration}
As described in [18], LSTMs suffer from predictive lag, wherein the model fails to react quickly to new clinical information. Previous work [18] demonstrated that an LSTM trained with input data perseveration (i.e. the input is replicated k times) responds with more pronounced changes in predictions when new measurements become available while maintaining overall performance relative to a standard LSTM. Because timely model responsiveness to acute clinical events is critical in determining the necessity of escalating support, input data perseveration was assessed as a training augmentation technique. 

\subsection{Transfer Learning}
Transfer learning (TL) is a technique of applying insights (e.g. data representations) that were previously learned from one problem to a new, related task [19-22].  It can be particularly beneficial when one task has significantly more training data than the other. To take advantage of the large CHLA PICU dataset, models using transfer learning techniques were also considered.  LSTM-based RNNs using the same input variables as those for the HFNC task were trained on over 9000 CHLA PICU episodes to predict ICU mortality [23]. The first layer of one of these mortality models was then used as the first layer of some of the LSTM-based HFNC prediction models in Table 3.  

\subsection{Ensembling}
Ensemble methods combine multiple algorithms to achieve higher predictive performance than each component could obtain [24]. The predictions from each of the components are averaged to yield a single, final prediction. Here, different seed values were used to generate multiple LSTM-based models, with each seed value initializing a different set of the pseudo-random starting weights for a particular model. Different seed values lead to slightly different models. Different seeds were used to train the mortality models used for Transfer Learning and to train the final LSTM models on HFNC-specific data. 

\subsection{HFNC Models}
Eight models were developed: a 14-variable Logistic Regression (LR-14) using variables previously identified as risk factors for HFNC failure [8, 11-13], a 517-variable Logistic Regression (LR-517), a standard LSTM, an LSTM with Input Perseveration (LSTM+3xPers, where “3” indicates the number of replications described in 24]), an LSTM with Transfer Learning (LSTM+TL), an LSTM with both Input Perseveration and Transfer Learning (LSTM+3xPers+TL), a simple ensemble of LSTMs with Input Perseveration and Transfer Learning (Simple-EN-LSTM+3xPers+TL), and an ensemble of ensembles of LSTMs with Input Perseveration and Transfer Learning (Multi-EN-LSTM+3xPers+TL). All models were trained to generate a prediction every time new measurements became available within the HFNC Period.

Table 3 describes the parameters of all models. Each model was developed on the training set to maximize the average of the validation set area under the receiver operating characteristic curves (AUROCs) measured hourly from 0 to 14 hours into the HFNC period. This window was selected to prioritize the most clinically impactful period of time.

\begin{table}[]
\caption{Details of the eight models considered.}
\label{tab:HFNC_Hyperparams}
\centering
\begin{tabular}{|l|l|}
\hline
\textbf{Model} & \textbf{Hyperparameters} \\ \hline
\begin{tabular}[c]{@{}l@{}}14 Variable Logistic Regression \\ (LR-14)\end{tabular} & \begin{tabular}[c]{@{}l@{}}Regularizer: 7.50e-1\\ Regularization: elasticnet (ratio   = 0.5)\end{tabular} \\ \hline
\begin{tabular}[c]{@{}l@{}}517 Variable Logistic Regression \\ (LR-517)\end{tabular} & \begin{tabular}[c]{@{}l@{}}Regularizer: 1.15e-3\\  Regularization: elasticnet (ratio   = 0.2)\end{tabular} \\ \hline
LSTM & \begin{tabular}[c]{@{}l@{}}Layers: 3\\   Number of Hidden Units: {[}128, 256, 128{]}\\  Batch Size: 12\\ Initial Learning Rate: 9.6e-4  \\ Patience: 10\\ Reduce Rate: 0.9\\ Number of Rate Reductions: 8\\  Loss Function: binary   cross-entropy\\ Optimizer: rmsprop\\  Dropout: 0.35\\ Recurrent Dropout: 0.2\\  Regularizer: 1e-4\\  Output Activation: sigmoid\end{tabular} \\ \hline
\begin{tabular}[c]{@{}l@{}}LSTM w/ 3-Times Input   Perseveration\\   (LSTM+3xPers)\end{tabular} & (same as LSTM) \\ \hline
\begin{tabular}[c]{@{}l@{}}LSTM w/ Transfer Learning\\   (LSTM+TL)\end{tabular} & \begin{tabular}[c]{@{}l@{}}(same as LSTM)\\ Transfer Weights: 1st   hidden layer only\end{tabular} \\ \hline
\begin{tabular}[c]{@{}l@{}}LSTM w/ 3-Times Input Perseveration   and Transfer Learning\\  (LSTM+3xPers+TL)\end{tabular} & \begin{tabular}[c]{@{}l@{}}(same as LSTM)\\  Transfer Weights: 1st   hidden layer only\end{tabular} \\ \hline
\begin{tabular}[c]{@{}l@{}}Simple Ensemble of LSTM+3xPers+TL\\  (Simple-EN-LSTM+3xPers+TL)\end{tabular} & \begin{tabular}[c]{@{}l@{}}(same as LSTM)\\ Transfer Weights: 1st   hidden layer only\end{tabular} \\ \hline
\begin{tabular}[c]{@{}l@{}}Multi-Ensemble of LSTM+3xPers+TL\\  (Multi-EN-LSTM+3xPers+TL)\end{tabular} & \begin{tabular}[c]{@{}l@{}}(same as LSTM)\\  Transfer Weights: 1st   hidden layer only\end{tabular} \\ \hline
\end{tabular}
\end{table}

Figure 3 illustrates how the ensemble models were formed.  An LSTM mortality model was trained (Seed A), and its first layer was used as the first layer (TL weights) of a 3-layer LSTM with Input Perseveration. Layers 2 and 3 of this model were trained on the HFNC data five times (Seeds 1-5), resulting in five slightly different HFNC models whose predictions were averaged to generate the Simple-EN-LSTM+3xPers+TL model predictions. This process was repeated four times to generate the ensemble of ensembles model: four LSTM mortality models were trained (Seeds A-D), each providing a different set of TL weights. For each of these four sets of TL weights, five different seeds were used to train with the HFNC data, resulting in twenty models whose predictions were averaged together to generate the Multi-EN-LSTM+3xPers+TL model predictions.

\begin{figure}[!t]
    \centering
    \includegraphics[width=.8\linewidth]{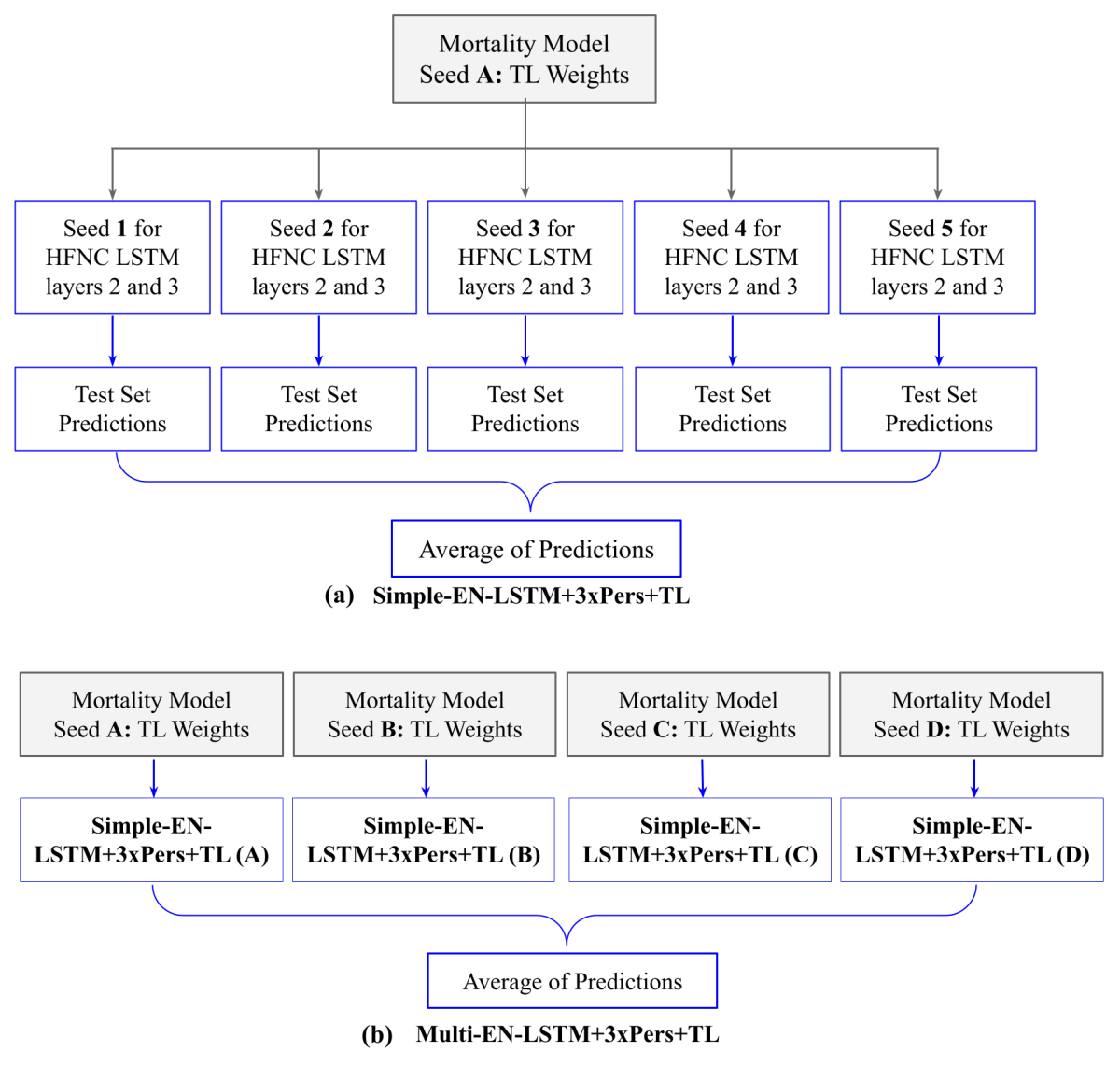}
    \caption{Forming the (a) Simple Ensemble and (b) Multi Ensemble models.}
    \label{fig:Ensemble}
\end{figure}

\subsection{Model Performance Assessment}
Model performance was assessed on the test set by evaluating the AUROC of predictions every 30 minutes within the 24-hour HFNC period. AUROC performance for the subset of patients with respiratory diagnoses was also compared every 30 minutes within the 24-hour HFNC Period. For all AUROC computations, failures or successes that already occurred before the time of evaluation were excluded. For example, any HFNC failures or successes that took place 4.5 hours or less into the HFNC Period were not considered in computing the 5-hour AUROC. Including these in the 5-hour AUROC calculation would artificially boost the result. This was followed for all time points of interest. As a result, the number of HFNC failures and successes in the test set steadily decreases from 0 to 24 hours into the HFNC Period.    

Additionally, ROC curves, Sensitivities, Specificities, Positive Predictive Values (PPV) and Negative Predictive Values (NPV) of predictions 2 hours after HFNC initiation were generated to evaluate model performance early in the HFNC Period, on both the entire test set cohort and the respiratory subcohort.
\section{Results}

\subsection{Cohort Characteristics}
Table 4 describes demographics and characteristics of the data, while Figure 4 shows the histogram (in cyan) and cumulative density (orange) for the time to HFNC failure for the entire dataset. 50\% of failures occurred within 7.6 hours, and 80\% occurred within 14.1 hours.

\begin{table}[]
\caption{Demographics and characteristics of the data partitions.}
\label{tab:HFNC_Demos}
\centering
\begin{tabular}{|ll|l|l|l|l|}
\hline
\multicolumn{2}{|l|}{\textbf{Characteristic}} & \textbf{Training  Set} & \textbf{Validation Set} & \textbf{Test  Set} & \textbf{Overall} \\ \hline
\multicolumn{2}{|l|}{Patients, n (\% died)} & 341 (5.9) & 138 (5.1) & 158 (3.2) & 637 (5.0) \\ \hline
\multicolumn{2}{|l|}{Episodes, n (\% died)} & 381 (5.2) & 151 (4.6) & 183 (2.7) & 715 (4.5) \\ \hline
\multicolumn{2}{|l|}{Total HFNC trials} & 455 & 173 & 206 & 834 \\ \hline
\multicolumn{2}{|l|}{\% HFNC Trials Died} & 4.6 & 5.8 & 3.4 & 4.6 \\ \hline
\multicolumn{2}{|l|}{\% HFNC Trials Failed} & 22.6 & 17.3 & 20.4 & 21.0 \\ \hline
\multicolumn{2}{|l|}{\% HFNC Trials Female} & 44.0 & 40.5 & 43.7 & 43.2 \\ \hline
\multicolumn{2}{|l|}{\% HFNC Trials with Respiratory   Primary Diagnosis} & 73.2 & 66.5 & 68.4 & 70.6 \\ \hline
\multicolumn{2}{|l|}{\textbf{PRISM III Score}} &  &  &  &  \\ \hline
\multicolumn{1}{|l|}{} & Mean (SD) & 4.4 (5.3) & 3.6 (5.0) & 4.0 (5.0) & 4.2 (5.2) \\ \hline
\multicolumn{1}{|l|}{} & Median {[}IQR{]} & 3 {[}0,7{]} & 2 {[}0,5{]} & 2 {[}0,6{]} & 3 {[}0,6{]} \\ \hline
\multicolumn{2}{|l|}{\textbf{Age (years)}} &  &  &  &  \\ \hline
\multicolumn{1}{|l|}{} & Mean (SD) & 3.3  (4.6) & 3.1 (4.5) & 2.8 (3.7) & 3.1  (4.4) \\ \hline
\multicolumn{1}{|l|}{} & Median  {[}IQR{]} & 1.2 {[}0.4 3.4{]} & 1.2  {[}0.5,3.1{]} & 1.2 {[}0.5, 3.8{]} & 1.2 {[}0.4, 3.5{]} \\ \hline
\multicolumn{2}{|l|}{\textbf{Age group, years (\%)}} &  &  &  &  \\ \hline
\multicolumn{1}{|l|}{} & {[}0, 1) & 45.1 & 45.1 & 46.6 & 45.4 \\ \hline
\multicolumn{1}{|l|}{} & {[}1, 5) & 36.0 & 36.4 & 37.4 & 36.5 \\ \hline
\multicolumn{1}{|l|}{} & {[}5, 10) & 6.8 & 6.9 & 8.3 & 7.2 \\ \hline
\multicolumn{1}{|l|}{} & {[}10, 19) & 12.1 & 11.6 & 7.8 & 10.9 \\ \hline
\end{tabular}
\end{table}

\begin{figure}[!t]
    \centering
    \includegraphics[width=.6\linewidth]{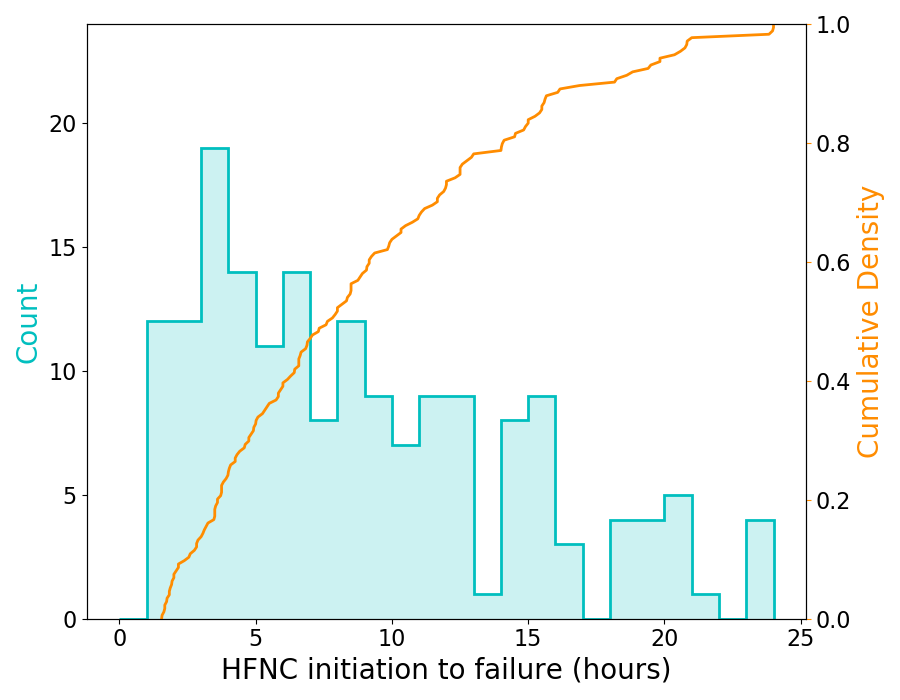}
    \caption{Distribution of Time to HFNC Failure.}
    \label{fig:TimeToFailure}
\end{figure}

\subsection{AUROC Across the First 24 Hours}
Figure 5 shows the eight models’ AUROCs in 30-minute increments within the 24-hour HFNC Period of all HFNC trials in the test set. Table A-7 in Multimedia Appendix 1 shows the number of remaining HFNC trials in the test cohort at various evaluation times. Table 5 presents the test set AUROC values associated with Figure 5 at several times of interest in the first 12 hours of the HFNC Period.  Table 6 presents the corresponding AUROCs in the respiratory cohort.

\begin{figure}[!t]
    \centering
    \includegraphics[width=.6\linewidth]{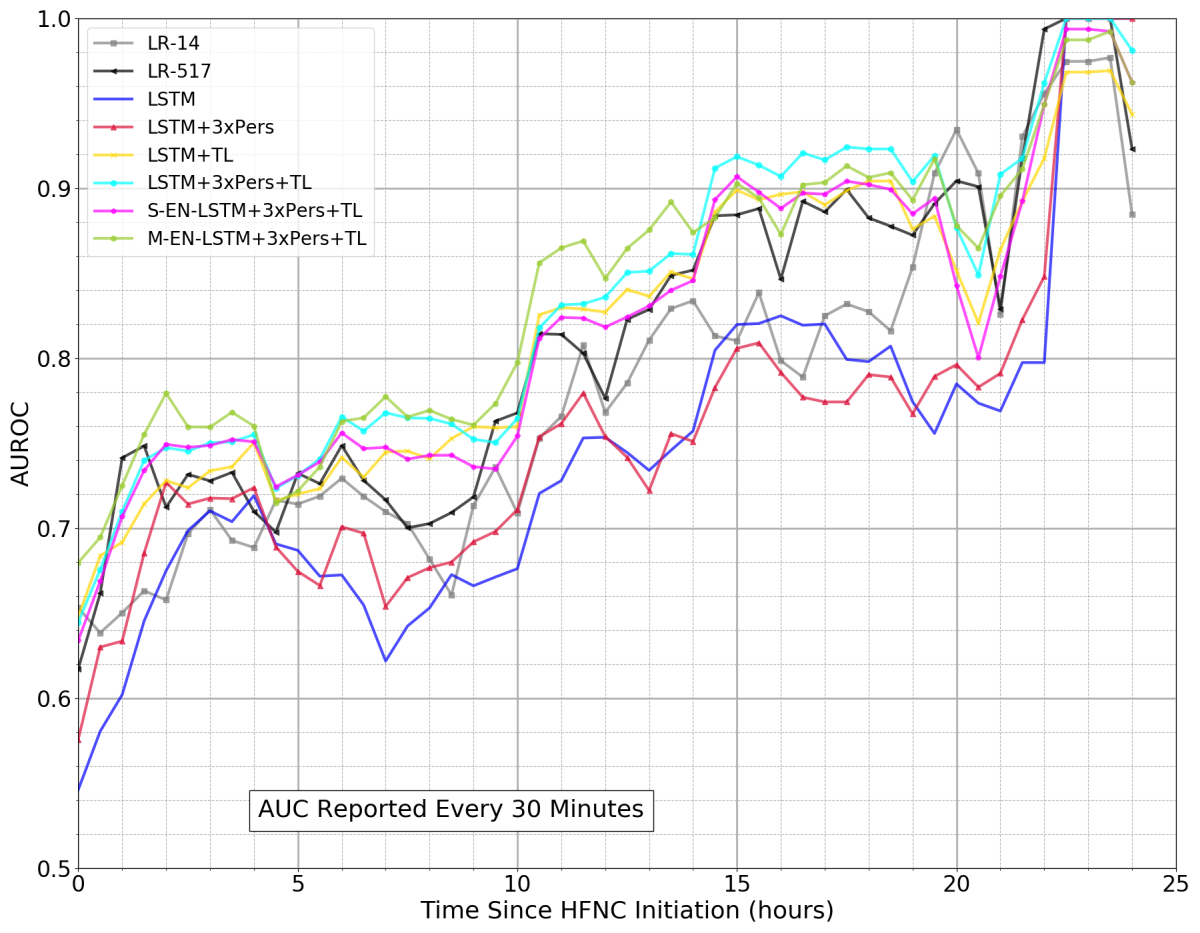}
    \caption{AUROCs of model predictions at different times on holdout test set.}
    \label{fig:nHour_AUCs}
\end{figure}

\begin{table}[]
\caption{Test set AUROCs of the eight models considered at various time points
following HFNC Initiation. The highest AUROC in a column is in bold font.}
\label{tab:HFNC_AUROC_time}
\centering
\begin{tabular}{|l|l|l|l|l|l|l|}
\hline
\textbf{Model} & \textbf{\begin{tabular}[c]{@{}l@{}}0-Hr \\ AUROC\end{tabular}} & \textbf{\begin{tabular}[c]{@{}l@{}}0.5-Hr \\ AUROC\end{tabular}} & \textbf{\begin{tabular}[c]{@{}l@{}}1-Hr \\ AUROC\end{tabular}} & \textbf{\begin{tabular}[c]{@{}l@{}}2-Hr \\ AUROC\end{tabular}} & \textbf{\begin{tabular}[c]{@{}l@{}}6-Hr \\ AUROC\end{tabular}} & \textbf{\begin{tabular}[c]{@{}l@{}}12-Hr \\ AUROC\end{tabular}} \\ \hline
\textbf{LR-14} & 0.65 & 0.64 & 0.65 & 0.66 & 0.73 & 0.77 \\ \hline
\textbf{LR-517} & 0.62 & \textbf{0.66} & 0.74 & 0.71 & 0.75 & 0.78 \\ \hline
\textbf{LSTM} & 0.55 & 0.58 & 0.60 & 0.68 & 0.67 & 0.75 \\ \hline
\textbf{LSTM+3xPers} & 0.58 & 0.63 & 0.63 & 0.73 & 0.70 & 0.75 \\ \hline
\textbf{LSTM+TL} & 0.65 & 0.68 & 0.69 & 0.73 & 0.74 & 0.83 \\ \hline
\textbf{LSTM+3xPers+TL} & 0.64 & 0.68 & 0.71 & 0.75 & \textbf{0.77} & 0.84 \\ \hline
\textbf{Simple-EN-LSTM+3xPers+TL} & 0.63 & 0.67 & 0.71 & 0.75 & 0.76 & 0.82 \\ \hline
\textbf{Multi-EN-LSTM+3xPers+TL} & \textbf{0.68} & \textbf{0.69} & 0.73 & \textbf{0.78} & 0.76 & \textbf{0.85} \\ \hline
\end{tabular}
\end{table}

\begin{table}[]
\caption{Test set AUROCs of the eight models considered at various time points 
following HFNC initiation for children with a respiratory diagnosis. The highest AUROC in a column is in bold font.}
\label{tab:HFNC_AUROC_time}
\centering
\begin{tabular}{|l|l|l|l|l|l|l|}
\hline
\textbf{Model} & \textbf{\begin{tabular}[c]{@{}l@{}}0-Hr  \\ AUROC\end{tabular}} & \textbf{\begin{tabular}[c]{@{}l@{}}0.5-Hr  \\ AUROC\end{tabular}} & \textbf{\begin{tabular}[c]{@{}l@{}}1-Hr  \\ AUROC\end{tabular}} & \textbf{\begin{tabular}[c]{@{}l@{}}2-Hr  \\ AUROC\end{tabular}} & \textbf{\begin{tabular}[c]{@{}l@{}}6-Hr \\ AUROC\end{tabular}} & \textbf{\begin{tabular}[c]{@{}l@{}}12-Hr  \\ AUROC\end{tabular}} \\ \hline
\textbf{LR-517} & 0.66 & 0.66 & \textbf{0.77} & 0.78 & \textbf{0.84} & 0.85 \\ \hline
\textbf{LSTM} & 0.53 & 0.61 & 0.63 & 0.70 & 0.76 & 0.82 \\ \hline
\textbf{Multi-EN-LSTM+3xPers+TL} & \textbf{0.70} & \textbf{0.70} & 0.74 & \textbf{0.82} & 0.83 & \textbf{0.89} \\ \hline
\end{tabular}
\end{table}

\subsection{2-Hour ROC and AUROC}
Figure 6 presents the test set 2-Hour ROC curves and AUROC for the eight models, showing predictive performance just two hours into the HFNC Period, while Figure 7 presents the same metrics corresponding to the respiratory cohort. 

\begin{figure}[!t]
    \centering
    \includegraphics[width=.6\linewidth]{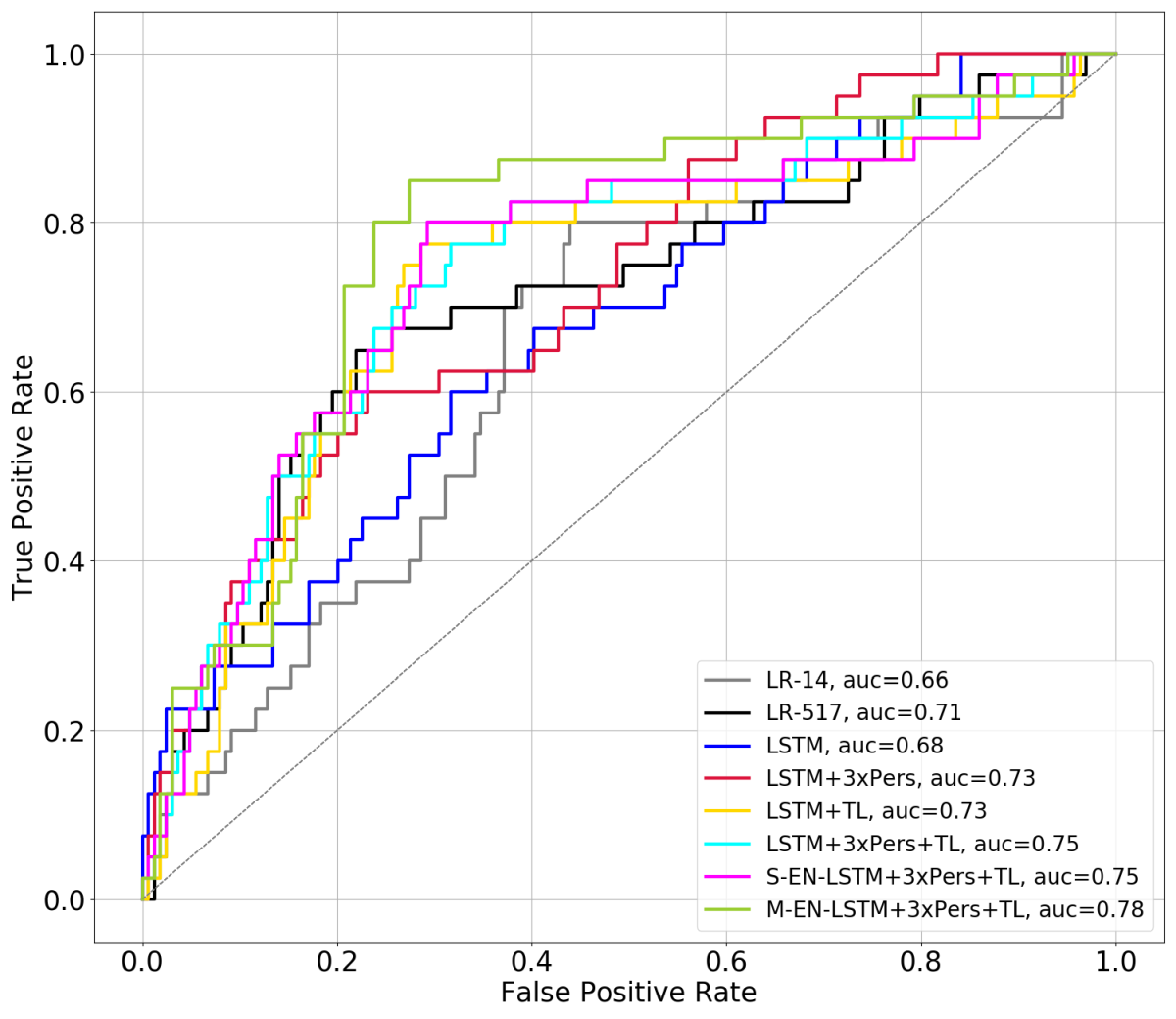}
    \caption{ROC Curves and AUROCs of 2-hour predictions on entire test set.}
    \label{fig:2Hour_AUCs}
\end{figure}

\begin{figure}[!t]
    \centering
    \includegraphics[width=.6\linewidth]{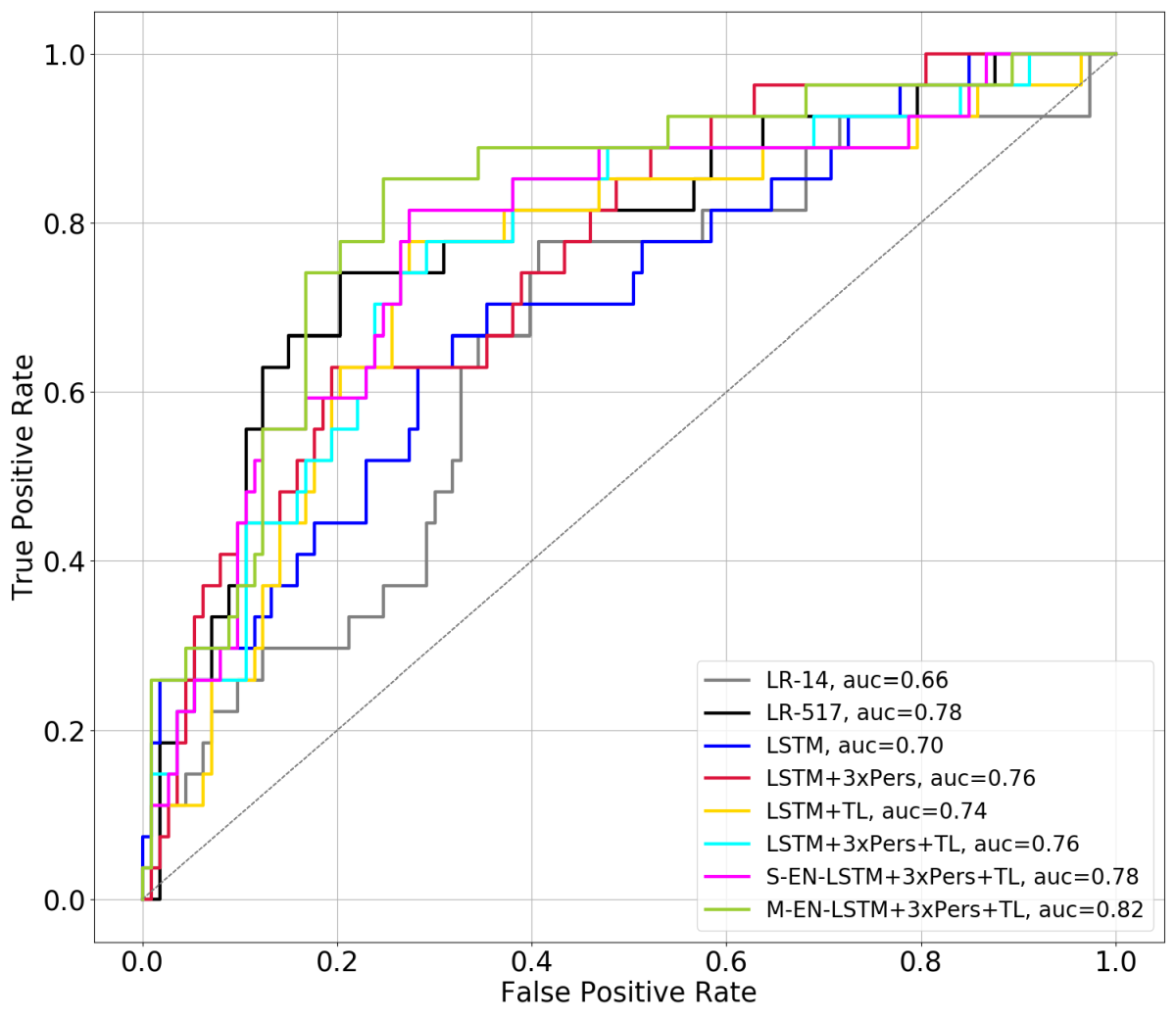}
    \caption{ROC Curves and AUROCs of 2-hour predictions on HFNC trials whose primary diagnosis is respiratory.}
    \label{fig:2Hour_AUCs_Resp}
\end{figure}

\subsection{PPV and NPV}
Table 7 shows the Specificity, Positive Predictive Value (PPV) and Negative Predictive Value (NPV) for the 2-Hour predictions of the Multi-EN-LSTM+3xPers+TL model that correspond to different values of Sensitivity.  At this 2-hour mark, 204 HFNC Periods remained (40 failures, 164 successes).  Tables A-8 through A-10 in Multimedia Appendix 1 show the comparison of these metrics across all models.

\begin{table}[]
\begin{tabular}{|l|l|l|l|}
\hline
\textbf{Sensitivity} & \textbf{Specificity} & \textbf{Positive   Predictive Value (PPV)} & \textbf{Negative   Predictive Value (NPV)} \\ \hline
0.10 & 0.982 & 0.571 & 0.817 \\ \hline
0.20 & 0.970 & 0.615 & 0.832 \\ \hline
0.30 & 0.927 & 0.500 & 0.844 \\ \hline
0.40 & 0.848 & 0.390 & 0.853 \\ \hline
0.50 & 0.835 & 0.426 & 0.873 \\ \hline
0.60 & 0.793 & 0.414 & 0.890 \\ \hline
0.70 & 0.793 & 0.452 & 0.915 \\ \hline
0.80 & 0.762 & 0.451 & 0.940 \\ \hline
0.90 & 0.463 & 0.290 & 0.950 \\ \hline
0.95 & 0.207 & 0.226 & 0.944 \\ \hline
1.00 & 0.049 & 0.204 & 1.000 \\ \hline
\end{tabular}
\end{table}
\section{Discussion}

\subsection{Primary Results}
The ability to predict reliably and in real time a child’s response to HFNC could help guide clinical differentiation among three groups: a) children most likely to do well on HFNC alone, b) children most likely to need higher-level support, and c) children whose likely HFNC outcome is unclear and who may require additional observation. Patients identified from the first group may require less clinical intervention and free up scarce ICU resources. Identifying children in the second group may enable clinicians to intervene more rapidly and provide adequate support to prevent decompensation. Due to clinical uncertainty, children in the third group may benefit from more careful and frequent observation with continuous prediction of the likelihood of failure.

The granular longitudinal data captured from children in the ICU presents a tremendous amount of information available for learning and developing tools to help differentiate children’s responses to numerous ICU interventions such as HFNC and including ventilation, ECMO, dialysis etc. Deep learning models, especially those with sequential processing capabilities such as LSTMs, have the potential to use rich time-dependent data in ways that more traditional ML models (e.g. Logistic Regression) cannot, but LSTMs may require sizable training data to construct generalizable models. The results from this study showed this to be the case: a standard, 3-layer LSTM was generally the worst performing model on the holdout test set. 

Transfer Learning was incorporated to address the issue of training LSTMs with insufficient data. The models with TL had the advantage of learning representations from more than 9000 PICU episodes, while the models without TL learned from just over 600 HFNC trials (about 500 episodes). The results demonstrate considerable gains from using TL and are consistent with theory [14]. Figure 5 and Table 5, in particular, highlight the significant and time-independent performance increase delivered by transfer learning in the LSTM models. 

Input Perseveration by itself (LSTM+3xPers) provided some performance boost relative to the standard LSTM, especially in the first 12 hours of the HFNC period in respiratory patients (Table 6). When combined with Transfer Learning (LSTM+3xPers+TL), it continued to provide additional, though slight, gains. As demonstrated in [13], LSTMs can suffer from a predictive lag phenomenon, wherein they fail to react rapidly to new data reflecting sudden clinical events and changes in patient status. In the context of HFNC usage and decisions about whether to escalate a child to higher levels of support, this predictive lag may be deleterious in a time constrained environment such as the PICU. 

Finally, the Ensemble models (Simple-EN-LSTM+3xPers+TL and Multi-EN-LSTM\_3xPers\_TL) were built to address another consequence of limited training data: the relatively high variability of any one particular realization of the model. This is a byproduct of randomly chosen initialization seeds used both to initialize LSTM weights and biases and for random dropout techniques used for regularization purposes). The ensemble methods provided consistently higher performance on the holdout test set compared to the non-ensemble models. The Ensemble models provided a slight performance boost over just a single LSTM+3xPers+TL model. Not surprisingly, the multi-ensembling of multiple models (both those used to generate Transfer Learning weights and those used to generate HFNC predictions) provided the best overall model (Multi-EN-LSTM+3xPers+TL).  

Model performance in patients with respiratory diagnoses is of interest because the pathophysiology of respiratory illness is particularly amenable to HFNC therapy [1-4]. Approximately 70\% of HFNC initiations in this cohort were in patients with respiratory illness. Table 6 shows that all models except LR-14 generally performed better in the respiratory group across time. Figure 7 shows that the best performing models in the overall cohort -- those that incorporated TL -- performed even better in the respiratory group after two hours of observation, demonstrating the TL models’ potential clinical impact. 

The 2-hour mark after HFNC initiation is an important clinical decision point because a child has had adequate time to adapt to HFNC, and the effects of treatment can be assessed. This motivated the additional analyses of 2-hour predictions shown in Figures 6-7 (ROC curves) and Table 7 (sensitivity, specificity, PPV, NPV at various decision thresholds). The Multi-EN-LSTM+3xPers+TL model had the highest AUROC. On this model’s ROC curve for the entire cohort, two operating points are of particular interest: the first corresponds to 95\% sensitivity (20\% specificity), and the second to 20\% sensitivity (97\% specificity). The first point can be used to identify children most likely to do well in HFNC (Group 1), while the second can identify those most likely to fail HFNC (Group 2).  Successfully identifying 20\% of Group 1 can reduce the observational burden, while identifying 20\% of Group 2 could lead clinicians to intervene earlier with an escalation to a higher level of O2 support, potentially improving outcome for these children [9,10]. Children for whom the model predictions fall between the two thresholds are in the third group: those whose HFNC outcome is unclear and who may benefit from more frequent observations. 

\subsection{Limitations}
This study had several limitations. First, it was based on just a single center retrospective cohort. Next, the target definition considered only the first 24 hours following HFNC initiation. Further work can refine the target to consider the subsequent 24 hours, regardless of how long the patient has already been on HFNC.

Finally, this study is limited by the exclusion of children suffering from apnea, making the predictive model’s applicability to such children unclear. Though less than ideal, this exclusion was deemed necessary since it is difficult to determine if escalation to BiPAP in these children is due to clinical necessity (i.e., true escalation) or just prophylactic caution (to guard against sleep apnea).

\section{Conclusion}

This study demonstrated the feasibility of applying advanced ML methodology to a complex and challenging clinical situation. This work demonstrated that clinically relevant models can be trained to predict the risk of escalation from HFNC within 24 hours of initiation of therapy could be obtained by using an LSTM with application of transfer learning and input perseveration to boost AUROC performance.

\section{Acknowledgements}
The authors would like to express their sincere gratitude to the Whittier Foundation for funding this work.

\section{Conflicts of Interest}
The authors have no actual or potential conflicts of interest in relation to the content of this study.

\section{Abbreviations}
AUROC: Area Under the Receiver Operating Curve \\
BiPAP: Bilevel Positive Airway Pressure \\
CHLA: Children’s Hospital Los Angeles \\
CPAP: Continuous Positive Airway Pressure \\ 
$FiO_2$: Fraction of Inspired Oxygen \\
FPR: False Positive Rate \\
HFNC: High Flow Nasal Cannula \\
IRB: Institutional Review Board \\
LR: Logistic Regressions \\
LSTM: Long Short-Term Memory \\ 
ML: Machine Learning \\
NIV: Non-Invasive Ventilation \\
$O_2$: Oxygen \\ 
RNN: Recurrent Neural Network \\
SF: Pulse Oximetric Saturation  \\
$SpO_2$: Oxygen Saturation \\
TPR: True Positive Rate \\
VPICU: Virtual Pediatric Intensive Care Unit




\bibliographystyle{unsrtnat}
\bibliography{references}  
1.	Kawaguchi A, Garros D, Joffe A, DeCaen A, Thomas NJ, Schibler A, Pons-Odena M, Udani S, Takeuchi M, Junior JC, Ramnarayan P. Variation in practice related to the use of high flow nasal cannula in critically ill children. Pediatric Critical Care Medicine; 2020;21(5):e228-35. [doi: 10.1097/PCC.0000000000002258] \\
2.	Habra B, Janahi IA, Dauleh H, Chandra P, Veten A. A comparison between high‐flow nasal cannula and noninvasive ventilation in the management of infants and young children with acute bronchiolitis in the PICU. Pediatric Pulmonology; 2020;55(2):455-461. [doi:10.1002/ppul.24553] \\
3.	Clayton JA, McKee B, Slain KN, Rotta AT, Shein SL. Outcomes of children with bronchiolitis treated with high-flow nasal cannula or noninvasive positive pressure ventilation. Pediatric Critical Care Medicine; 2019; 20(2):128-35. [doi: 10.1097/PCC.0000000000001798] \\
4.	Ramnarayan P, Schibler A. Glass half empty or half full? The story of high-flow nasal cannula therapy in critically ill children. Intensive Care Medicine; 2017; 43:246-249. [doi:10.1007/s00134-016-4663-2] \\
5.	Coletti KD, Bagdure DN, Walker LK, Remy KE, Custer JW. High-flow nasal cannula utilization in pediatric critical care. Respiratory Care; 2017; 62(8):1023-9. [doi:10.4187/respcare.05153] \\
6.	Mikalsen IB, Davis P, Øymar K. High flow nasal cannula in children: a literature review. Scandinavian Journal of Trauma, Resuscitation and Emergency Medicine; 2016;24(1):1-2. [doi:10.1186/s13049-016-0278-4]  \\
7.	Slain KN, Shein SL, Rotta AT. The use of high-flow nasal cannula in the pediatric emergency department. Jornal de Pediatria; 2017;93:36-45. [doi:10.1016/j.jped.2017.06.006] \\
8.	Kelly GS, Simon HK, Sturm JJ. High-flow nasal cannula use in children with respiratory distress in the emergency department: predicting the need for subsequent intubation. Pediatric Emergency Care; 2013;29(8):888-92. [doi:10.1097/PEC.0b013e31829e7f2f] \\
9.	Bauer PR, Gajic O, Nanchal R, Kashyap R, Martin-Loeches I, Sakr Y, Jakob SM, François B, Wittebole X, Wunderink RG, Vincent JL. Association between timing of intubation and outcome in critically ill patients: A secondary analysis of the ICON audit. Journal of critical care; 2017;42:1-5. [doi:10.1016/j.jcrc.2017.06.010] \\
10.	Kang BJ, Koh Y, Lim CM, Huh JW, Baek S, Han M, Seo HS, Suh HJ, Seo GJ, Kim EY, Hong SB. Failure of high-flow nasal cannula therapy may delay intubation and increase mortality. Intensive care medicine; 2015;41(4):623-32. [doi:10.1007/s00134-015-3693-5] \\
11.	Guillot C, Le Reun C, Behal H, et al. First-line treatment using high-flow nasal cannula for children with severe bronchiolitis: Applicability and risk factors for failure. Archives de Pédiatrie; 2018;25(3):213-218. [doi:10.1016/j.arcped.2018.01.003] \\
12.	Er A, Çağlar A, Akgül F, et al. Early predictors of unresponsiveness to high-flow nasal cannula therapy in a pediatric emergency department. Pediatric Pulmonology; 2018;53(6):809-815. [doi:10.1002/ppul.23981] \\
13.	Yurtseven A, Ulaş Saz E. The effectiveness of heated humidified high-flow nasal cannula in children with severe bacterial pneumonia in the emergency department. The Journal of Pediatric Research; 2020;7(1):71-76. [doi:10.4274/jpr.galenos.2019.15045] \\
14.	https://www.myvps.org \\
15.	Ho LV, Ledbetter D, Aczon M, Wetzel R. The dependence of machine learning on electronic medical record quality. AMIA Annu Symp Proc. 2017; (2017):883-891. [PMID: 29854155; PMCID: PMC5977633.] \\
16.	Imholz BP, Settels JJ, van der Meiracker AH, Wesseling KH, Wieling W. Non-invasive continuous finger blood pressure measurement during orthostatic stress compared to intra-arterial pressure. Cardiovascular Research; 1990;24(3):214-221. [doi:10.1093/cvr/24.3.214] \\
17.	Schulman CS, Staul LA. Standards for frequency of measurement and documentation of vital signs and physical assessments. Critical Care Nurse; 2010;30(3):74-76. [doi:10.4037/ccn2010406] \\
18.	Ledbetter D, Laksana E, Aczon M, Wetzel R. Improving recurrent neural network responsiveness to acute clinical events. https://arxiv.org/abs/2007.14520 \\
19.	Tan C, Sun F, Kong T, Zhang W, Yang C, Liu C. A survey on deep transfer learning. The 27th International Conference on Artificial Neural Networks (ICANN 2018). https://arxiv.org/abs/1808.01974 \\
20.	Silver DL, Bennett KP. Guest editor’s introduction: special issue on inductive transfer learning. Machine Learning; 2008;73(3):215-20. [doi: 10.1007/s10994-008-5087-1] \\
21.	Weiss, K., Khoshgoftaar, T.M. and Wang, D. A survey of transfer learning. J Big Data 3, 9 (2016). [https://doi.org/10.1186/s40537-016-0043-6] \\
22.	Niu S, Liu Y, Wang J, Song H. A decade survey of transfer learning (2010–2020). IEEE Transactions on Artificial Intelligence; 2020;1(2):151-66. \\
23.	Aczon M, Ledbetter D, Laksana E, Ho L, Wetzel R. Continuous Prediction of Mortality in the PICU: A Recurrent Neural Network Model in a Single-Center Dataset. Pediatric Critical Care Medicine; 2021;22(6):519-529. [doi: 10.1097/PCC.0000000000002682] \\
24.	Ganaie MA, Hu M. Ensemble deep learning: A review. arXiv preprint arXiv:2104.02395. 2021 Apr 6.

\end{document}